# Clustering using Vector Membership: An Extension of the Fuzzy C-Means Algorithm

Srinjoy Ganguly[1], Digbalay Bose[2], Amit Konar[3]

[1,2,3] Department of Electronics & Telecommunication Engineering, Jadavpur University, Kolkata, India

[1]srinjoy_ganguly92@hotmail.com, [2]digbose92@gmail.com, [3]konaramit@yahoo.com

*Abstract*— Clustering is an important facet of explorative data mining and finds extensive use in several fields. In this paper, we propose an extension of the classical Fuzzy C-Means clustering algorithm. The proposed algorithm, abbreviated as VFC, adopts a multi-dimensional membership vector for each data point instead of the traditional, scalar membership value defined in the original algorithm. The membership vector for each point is obtained by considering each feature of that point separately and obtaining individual membership values for the same. We also propose an algorithm to efficiently allocate the initial cluster centers close to the actual centers, so as to facilitate rapid convergence. Further, we propose a scheme to achieve crisp clustering using the VFC algorithm. The proposed, novel clustering scheme has been tested on two standard data sets in order to analyze its performance. We also examine the efficacy of the proposed scheme by analyzing its performance on image segmentation examples and comparing it with the classical Fuzzy C-means clustering algorithm.

## I. INTRODUCTION

Clustering [1] is one of the most powerful tools of data mining, which primarily deals with the task of organization of data into specific groups, whose characteristics are unknown. It primarily involves dividing a set of patterns into clusters or groups, such that objects in one cluster are very similar to each other while objects in different clusters are quite distinct. Clustering unlike classification, involves grouping of unlabelled patterns into meaningful clusters, where the grouping is solely driven by the data. The clustering algorithms which are primarily used are divided into multiple categories, depending upon the methodologies adopted [2], i.e. agglomerative or divisive, hard or fuzzy etc. Hard clustering algorithm primarily segregates the data group into clusters by assigning a particular cluster to each member, whereas a fuzzy clustering method assigns degrees of membership in several clusters to each input data. The idea of fuzzy clustering has gained widespread recognition, since the introduction of the idea of fuzzy sets by Zadeh [3-4]. The most widely used among all the fuzzy clustering models is the fuzzy C-means clustering, proposed by Dunn [5] and further studied by Bezdek [6]. The basic idea of fuzzy C means has inspired multiple new variants including fuzzy k-neighbourhood algorithm [7], entropy based clustering [8], potential based clustering [9], possibilistic fuzzy C-means clustering (PFCM) [10], etc.

In this paper, we wish to introduce a novel **V**ector **F**uzzy **C**-means clustering algorithm (VFC), an extension of the fuzzy C-means clustering algorithm which utilizes a vector membership function by considering each dimension (feature) of a data point separately. Therefore, each feature is processed separately by the VFC, instead of clubbing all of them into one as per the original approach, so as to consider the diversity of each feature separately for more accurate clustering. An efficient strategy for initialization of the initial cluster centers has been also discussed which ensures sufficient knowledge is available initially for determination of the optimal solution. Further, we have proposed a novel scheme to facilitate crisp clustering utilizing the fuzzy membership vectors computed by the proposed VFC algorithm. The efficacy of the proposed scheme is verified by analyzing its performance when used to cluster two standard data-sets, namely the Wisconsin Breast Cancer Data-set (WBCD) and the Iris Data-set. The performance of the stated protocol vis-à-vis other relevant fuzzy clustering protocols is compared by virtue of their performances in an image segmentation task. A list of abbreviations and acronyms used frequently throughout the paper is listed in Table 1. The rest of the paper has been organized as follows. Section II discusses the classical fuzzy C-means (FCM) clustering algorithm in brief. The VFC algorithm is detailed in Section III. The experimental results and related discussions are presented in Section IV. Section V concludes the paper with a brief analysis of the experimental results and scope of future work.

TABLE I. LIST OF ABBREVIATIONS AND ACRONYMS

| | |
|---|---|
| $C$ | Number of clusters |
| $X$ | Set of data points |
| $D$ | Number of dimensions in each data point |
| $N$ | Number of data points |
| $V$ | Set of cluster centers |
| $v_j$ | The $j^{th}$ cluster centre |
| $v_{jk}$ | The $k^{th}$ dimension of the $j^{th}$ cluster centre |
| $U$ | Membership Matrix (Dimension: $N * C$) |
| $u_{ij}$ | Degree of membership of the $i^{th}$ data point in the $j^{th}$ cluster ($0 \leq u_{ij} \leq 1$) |
| $u_{ijk}$ | Degree of membership of the $k^{th}$ dimension of $i^{th}$ data point in the $j^{th}$ cluster ($0 \leq u_{ijk} \leq 1$) |
| $d_{ij}$ | Euclidean Distance between the $i^{th}$ and the $j^{th}$ cluster centers |
| $m$ | Fuzziness Index ($1 \leq m < \infty$) |
| $J$ | Objective Function to be minimized |

## II. CLASSICAL FCM CLUSTERING

Fuzzy C-Means (FCM) clustering is one of the most popular fuzzy clustering techniques, where each data element may be the member of multiple clusters at the same time. FCM involves an iterative procedure where the dissimilarity measure, given in terms of Euclidean distance, is minimized by updating the cluster centers and membership values according to the formulae:

$$\mu_{ij} = \frac{1}{\sum_{k=1}^{C}\left(\frac{d_{ij}}{d_{ik}}\right)^{\frac{2}{m-1}}} \quad (1)$$

$$v_j = \frac{\sum_{i=1}^{N}(\mu_{ij})^m x_i}{\sum_{i=1}^{N}(\mu_{ij})^m}, \forall j = 1, 2, ..., C \quad (2)$$

The main objective is to minimize the objective function:

$$J(U,V) = \sum_{i=1}^{N}\sum_{j=1}^{C}(\mu_{ij})^m \| x_i - v_{ij} \|^2 \quad (3)$$

This iterative process repeats itself till a stopping criterion as deemed suitable by the user is met.

## III. THE VFC ALGORITHM

### (A) Multi-dimensional Fuzzy Membership Vector

As per the traditional Fuzzy C-Means clustering approach, each data element has a certain degree of membership $0 \leq u_{ij} \leq 1$, where is a scalar quantity that represents the degree of membership of the data element $x_k$ in the $i^{th}$ partitioning fuzzy subset(cluster) of the data-set $X$. In the proposed VFC algorithm, we have proposed a novel vector membership scheme: $u_i = [u_{ij1}, u_{ij2}, ..., u_{ijD}]$, where the membership of each dimension (feature) of the data element $x_k$ in a fuzzy subset (say, the $i^{th}$ cluster) is taken into consideration. We have adopted the said scheme to ensure more efficient clustering as each dimension of each D – dimensional data element is processed separately, hence, replacing the scalar membership function that merely computes an aggregate quantity.

### (B) Refined Initial Cluster Centers

Another novel feature that we wish to propose is a novel algorithm for efficient initialization of initial cluster centers that may be close to the actual optimal cluster centers, hereby facilitating a rapid convergence to the optimal solution. In this regard, we intend to throw some light on a metric that we have utilized for generation of efficient initial cluster centers: Degree of Scattering(s). It represents the extent to which the data corresponding to a particular dimension (or, feature) of all the data elements is scattered. Hence, for the $r^{th}$ dimension, the value of $s_r$ is given by:

$$s_r = \frac{\left(X_r^{\max} - X_r^{\min}\right)}{\sigma_{X_r}} \quad (4)$$

Where $X_r$ refers to the set of values represented by the $r^{th}$ dimension of the data-set X and σ represents the standard deviation of set. The greater will be the value of $s_r$, the more will the data values corresponding to the $r^{th}$ dimension said to be scattered. The proposed algorithm has been represented in Fig 1.

```
BEGIN  procedure
The Degree of Scattering for each dimension s_r is computed.
The Centroid corresponding to each dimension C_d is
computed.
// N – Number of data elements in the data set.
// D – Number of dimensions in the data set.
FOR i = 1; i ≤ N
   FOR j = 1; j ≤ D

     wd_i = wd_i + s_j × | X_ij − C_j |   // wd_i – Weighted distance
of the i^th data vector from the centroid.
       j = j +1
     END j – loop
 i = i +1
END i – loop

Rearrange the data vectors based on the increasing values of
their respective weighted distance values.
Divide the entire rearranged data set into C clusters.
Find the centroid of each such cluster.
The obtained centroids are the required initial Cluster
Centers.
END  procedure
```

**Fig 1. Cluster Center Initialization Scheme**

*(C) Modified Objective and Update Functions*

As the proposed VFC considers a D – dimensional membership function, the new definition of the objective function that is to be minimized is:

$$J(U,V;X) = \sum_{i=1}^{N}\sum_{j=1}^{C}\sum_{k=1}^{D} u_{ijk} \times \| x_{ik} - v_{jk} \|^2 \tag{5}$$

The corresponding update functions are:

$$u_{ijk} = \frac{1}{\sum_{n=1}^{C}\left(\frac{\| x_{ik} - v_{jk} \|}{\| x_{ik} - v_{nk} \|}\right)^{\frac{2}{m-1}}} \tag{6}$$

$$v_{jk} = \frac{\sum_{i=1}^{N}(u_{ijk}^{m} \times x_{ik})}{\sum_{i=1}^{N} u_{ijk}^{m}} \tag{7}$$

*(D) Stopping Criteria*

There are essentially two ways in which stopping criteria for the fuzzy C-means clustering algorithm can be defined. It can either stop if $\max_{ijk}(u_{ijk}^{(g+1)} - u_{ijk}^{g}) \leq \epsilon$ (where, $0 \leq \epsilon \leq 1$) or the number of iterations reach a certain maximum value $I_{max}$. We have chosen the latter as the stopping criterion for the proposed algorithm.

*(E) Scheme for the overall VFC Algorithm*

The scheme for the overall VFC algorithm has been represented in Fig. 2 . We have also proposed a scheme wherein we can obtain crisp clustering by means of the proposed VFC algorithm. The scheme has been represented in Fig. 3 .

```
BEGIN  VFC
Initialize V_inital = v_ij as described in Section III-(B).
Calculate U^(0) = u_ijk from V & X  as described in
Section III-(C).
 I = 0.
 while ( I ≤ I_max )
Evaluate the objective function J as described in
Section III-(C).
At the k^th step, calculate the new Cluster Centre
vectors V_new as described in Section III-(C).
Update the fuzzy membership matrix U  as described in
Section III-(C).
 I = I +1.
 end – while
END  VFC
```

**Fig 2. VFC Algorithm**

```
BEGIN VFC
Initialize V_inital = v_ij as described in Section III-(B).
Calculate U^(0) = u_ijk from V & X as described in
Section III-(C).
I = 0.
while ( I ≤ I_max )
Evaluate the objective function J as described in
Section III-(C).
At the k^th step, calculate the new Cluster Centre
vectors V_new as described in Section III-(C).
Update the fuzzy membership matrix U as described in
Section III-(C).
I = I +1.
end − while
END VFC
```

**Fig 3. Crisp VFC clustering scheme**

## IV. EXPERIMENTAL RESULTS

To test the effectiveness of our proposed algorithm, we have performed tests on certain standard data sets and included a test case for the purpose of image segmentation. In order to perform the tests, the proposed VFC algorithm is coded in MATLAB and the experiments are executed on a Pentium P-IV 3.0GHz PC with 512MB memory. The Experimental results section is divided into two subsections A and B, where subsections A and B detail the results associated with standard data sets and image segmentation test case, respectively.

*(A) Results obtained on the Standard Data-sets*

In this section certain standard data sets have been considered. The data-sets on which we have tested the algorithm are:

- The IRIS dataset [11], which consists of three classes of 50 instances each, where each class refers to a type of the iris plant. One class is linearly separable from the others, while this doesn't hold true for the remaining two.
- The Wisconsin Breast Cancer dataset [12] which is a multivariate data set comprising of 569 instances and possessing 32 attributes.

The effectiveness of our algorithm in case of Wisconsin Breast Cancer data set is highlighted in the Fig 1, where the normalized value of the objective function given by equation No. 5, is plotted against the number of iterations. The number of clusters considered were 4 and the degree of fuzziness i.e. m, was fixed at 5.

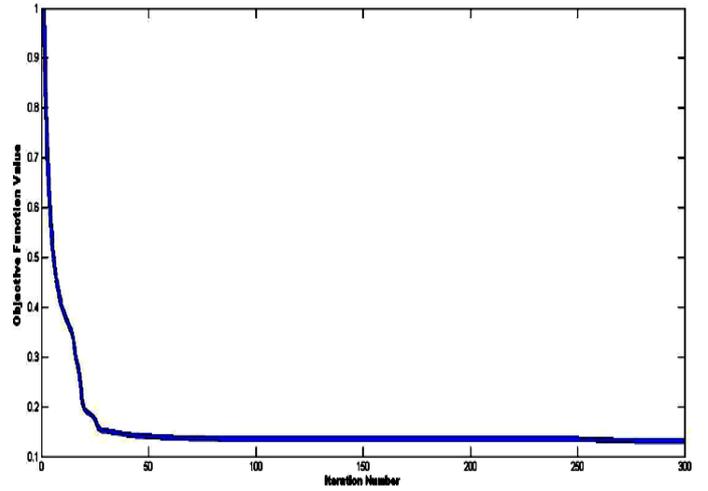

**Fig 4. A plot of the normalized objective function value VS the number of iterations for the WBCD dataset**

For the Iris data set, we present a 3-D plot of the first three features, since the fourth feature, the petal width, is highly correlated with the third measurement, the petal length. The cluster centers have been marked with black dots and the respective clusters are marked with different colours. The number of clusters were 4 and the degree of fuzziness i.e. m was fixed at 5. For the purpose of plotting the first three features the crisp clustering segment of VFC, as detailed in the previous section, is invoked to determine the appropriate cluster corresponding to each data point.

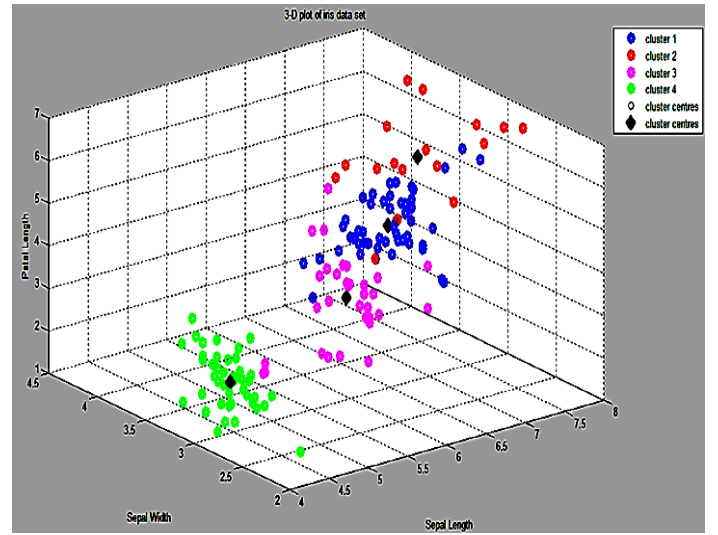

**Fig 5. A three-dimensional plot of the first three features of the iris data set. The points coloured blue, red, pink and green are the members of clusters 1, 2, 3 and 4, respectively. The cluster centers are marked by bold, black dots.**

*(B) Image Segmentation Results*

The processing of an image is a computationally intensive operation, due to which the idea of image segmentation [13], i.e. the subdivision of an image into its constituent regions or objects, plays a crucial role. The segmentation process terminates only after the objects of interest have been separated.

Since clustering is primarily a process utilized for classification of patterns and objects, it has emerged as a popular method in the domain of image segmentation. Depending upon the nature of the clustering algorithm (crisp or fuzzy) the segmentation can be either crisp or soft [14-16]. The fuzzy clustering algorithms tend to perform better in comparison with the crisp variants, due to their ability to retain more information, with the Fuzzy C Means algorithm being widely used for segmentation purposes.

In this section we investigate the performance of the VFC algorithm on the segmentation of the popular image called Lena, where the intensities of the image pixels are divided into two clusters. Depending upon the membership values of the pixels associated with the clusters, an intensity value of 0 is given to a pixel having membership value greater than or equal to 0.5 and a value of 1 to the pixel otherwise. The above mentioned intensity values are used for constructing the segmented image. The original Lena image used for image segmentation is shown in Fig 6. Here the performance of our proposed algorithm VFC is compared with the classical FCM (Fuzzy C Means). Both FCM and VFC have an exponential weighting factor m, whose impact on their individual performances is analysed by varying the value from 1.1 to 4.5.

The performance of FCM does not change with the variation in m, as depicted in Fig 7. However in case of VFC , when m=4.5 , better performance is obtained in terms of greater distinction in the nose and upper lip region, as shown in Fig 8.

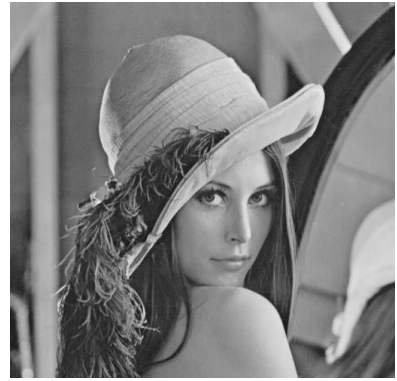

**Fig 6. Original Lena image (512 x512) used for image segmentation**

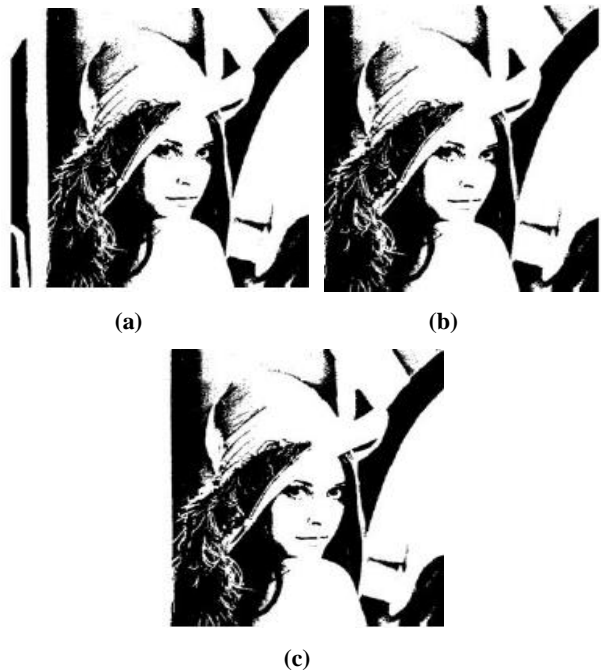

**(a)**           **(b)**

**(c)**

**Fig 7. Results of image segmentation by using the FCM algorithm (a)** *m=1.1* **(b)** *m=2.5* **(c)** *m= 4.5*

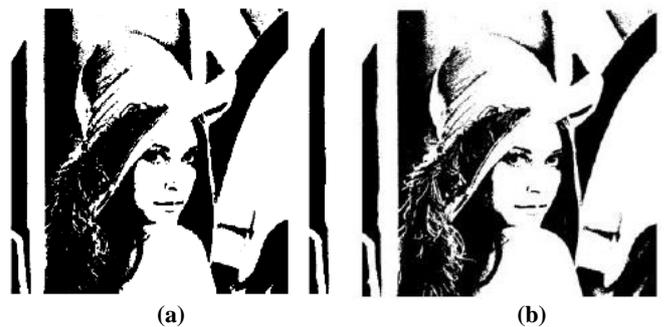

**(a)**           **(b)**

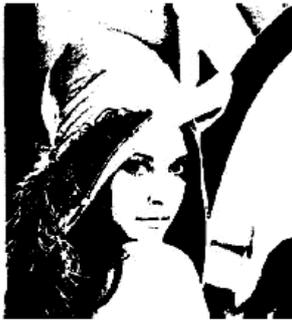

(c)

**Fig 8. Results of image segmentation by using the VFC algorithm (a)** *m=1.1* **(b)** *m=2.5* **(c)** *m= 4.5*

## V. CONCLUSION

In this paper, we proposed a novel fuzzy c-means clustering scheme that adopts a vectorized membership function, thereby replacing the traditional scalar membership value. Accordingly all the features of the clustering algorithm, be it the objective function or the membership & cluster centre update functions, have been modified accordingly. This proves to be beneficial as each attribute is processed individually, independent of the other attributes. This ensures that the proposed VFC is robust to noise & outliers, and can tolerate clusters that are unequal in size. Another important feature of the proposed algorithm is that it doesn't initialize the membership functions randomly. Rather, it determines the set of initial cluster centers as per a definite rule, after which it computes the corresponding membership functions. This facilitates convergence and ensures that the overall optimization process is speeded up. To analyze its performance, we have performed the tests on several datasets that include the famous Iris dataset and also the Wisconsin Breast Cancer data-set. We have also considered a test case for image segmentation task i.e. the commonly used image Lena and compared the results of our proposed algorithm with the classical FCM .We have also explained in brief how the algorithm can be also used for crisp clustering. Furthermore, this algorithm may be modified accordingly to find use in various areas such as bioinformatics, medicine, economics, etc.

## REFERENCES

[1] G.Gan, C.Ma, J.Wu," Data Clustering,Theory ,Algorithms and Applications", ASA-SIAM Series on Statistics and Applied Probability, SIAM, Philadelphia, ASA, Alexandria

[2] A.K.Jain, M.N.Murthy, P.J.Flynn, "Data Clustering: A Review", ACM Computing Surveys, Vol. 31, No. 3, September 1999

[3] L.A. Zadeh, Fuzzy sets, Inf. Control 8 (1965) 338–353

[4] M.S. Yang, "A survey of fuzzy clustering, Mathematical Computing and Modelling" 18 (1993) 1–16.

[5] J.C. Dunn, "A fuzzy relative of the ISODATA process and its use in detecting compact, well-separated clusters", J. Cybernet. 3 (1974) 32–57

[6] J.C. Bezdek, "Pattern Recognition with Fuzzy Objective Function Algorithms", Plenum, New York, 1981.

[7] J.C. Bezdek, "Fuzzy Mathematics in pattern classification", Ph.D Thesis, Applied Mathematics Centre, Cornell University, Ithaca,1973

[8] J.Yao ,M.Dash, S.T.Tan, H.Liu ,"Entropy based fuzzy clustering and fuzzy modeling", Fuzzy sets and Systems, 113, pp,381-388,2000

[9] S.L. Chiu, " Fuzzy model identification based on cluster estimation", Journal of Intelligent Fuzzy Systems, 2, pp 267-278, 1994

[10] N.R.Pal, K.Pal, J.M.Keller, J.C. Bezdek, "A Possibilistic Fuzzy c-Means Clustering Algorithm", IEEE Transactions on Fuzzy Systems, vol. 13, no. 4, August 2005

[11] E. Anderson, "The irises of the GASPE peninsula," in Bull. Amer. Iris Soc., vol. 59, 1935, pp. 2–5.

[12] W.H. Wolberg, W.N. Street, D.M. Heisey, and O.L. Mangasarian, "Computerized breast cancer diagnosis and prognosis from fine needle aspirates" Archives of Surgery 1995;130:511-516

[13] S. K. Pal and N. R. Pal, "A Review on Image Segmentation Techniques",Pattern Recognition, Vol. 26, No. 9, pp. 1277-1294, 1993

[14] G.B.Colemann, Harry C Andrews, "Image segmentation by clustering," *Proceedings of the IEEE* , vol.67, no.5, pp.773,785, May 1979

[15] Y Yang, S Huang, "Image clustering by Fuzzy C Means algorithm with a novel penalty term", Computing and Informatics, vol 26, 2007, 17-31.

[16] Zhu Feng; Song Yuqing; Chen Jianme, "Fuzzy C-Means Clustering for Image Segmentation Using the Adaptive Spatially Median Neighborhood Information," Chinese Conference on Pattern Recognition (CCPR), 2010, pp.1-5, 21-23 Oct. 2010

[17] M.R Rezae, Van der Zwet, P. M J; B.P.F Lelieveldt,.; R J van der Geest, ; J H C Reiber, "A multiresolution image segmentation technique based on pyramidal segmentation and fuzzy clustering," IEEE Transactions on Image Processing, vol.9, no.7, pp.1238-1248, July 2000 doi: 10.1109/83.847836

[18] M Tabakov, "A Fuzzy Clustering Technique for Medical Image Segmentation", 2006 International Symposium on Evolving Fuzzy Systems, pp 118-122, September 2006